\documentclass{article}

\usepackage{arxiv}
\usepackage[american]{babel}
\usepackage[utf8]{inputenc} % allow utf-8 input
\usepackage[T1]{fontenc}    % use 8-bit T1 fonts

\usepackage{hyperref}       % hyperlinks
\usepackage{url}            % simple URL typesetting
\usepackage{booktabs}       % professional-quality tables
\usepackage{amsfonts}
\usepackage{amsmath}
\usepackage{amssymb}        % blackboard math symbols
\usepackage{nicefrac}       % compact symbols for 1/2, etc.
\usepackage{microtype}      % microtypography
\usepackage{fullpage}
\usepackage{algorithmicx}
\usepackage{algorithm}
\usepackage{algpseudocode}
\usepackage{amsthm}
\usepackage{bm}
\usepackage{caption}
\usepackage{subcaption}
\DeclareMathAlphabet{\mathcal}{OMS}{cmsy}{m}{n}

\newtheorem{theorem}{Theorem}

\newtheorem{assump}{Assumption}
\newtheorem{remark}{Remark}

\newcommand{\x}{\mathbf x} %x-vector
\newcommand{\z}{\mathbf z} %z-vector
\renewcommand{\v}{\mathbf v}  %v-vector
\newcommand{\yh}{\hat y} %estimated y

\newcommand{\N}{\mathbb{N}} %natural numbers
\newcommand{\R}{\mathbb{R}} %real numbers
\renewcommand{\S}{\mathcal S} %Parameter space
\newcommand{\C}{\mathcal C} %Set of continuous functions
\newcommand{\X}{\mathcal X} %Input space
\newcommand{\Y}{\mathcal Y} %output space
\newcommand{\PP}{\mathcal P} %space of prob measures

\newcommand{\dett}{\operatorname{det}} %determinant 
\newcommand{\grad}{\operatorname{grad}} 
\newcommand{\divv}{\operatorname{div}} 
\newcommand{\normal}{\operatorname{\mathcal{N}}} %normal distribution

\usepackage{natbib} % has a nice set of citation styles and commands
\usepackage{mathtools} % amsmath with fixes and additions
\usepackage{booktabs} % commands to create good-looking tables
\usepackage{tikz} % nice language for creating drawings and diagrams

\title{Stochastic Mirror Descent in Average Ensemble Models}

\author{
  Taylan Kargin \\
  Department of Electrical Engineering\\
  California Institute of Technology\\
  Pasadena, CA 91125  \\
  \And
  Fariborz Salehi \\
  Google\\
  Seattle, WA 98103 \\
  \And
  Babak Hassibi \\
  Department of Electrical Engineering\\
  California Institute of Technology\\
  Pasadena, CA 91125 \\
}

\begin{document}
\maketitle

\begin{abstract}
The stochastic mirror descent (SMD) algorithm is a general class of training algorithms, which includes the celebrated stochastic gradient descent (SGD), as a special case. 
It utilizes a mirror potential to influence the implicit bias of the training algorithm. In this paper we explore the performance of the SMD iterates on mean-field ensemble models. Our results generalize earlier ones obtained for SGD on such models. 
The evolution of the distribution of parameters is mapped to a continuous time process in the space of probability distributions. Our main result gives a nonlinear partial differential equation to which the continuous time process converges in the asymptotic regime of large networks. The impact of the mirror potential appears through a multiplicative term that is equal to the inverse of its Hessian and which can be interpreted as defining a gradient flow over an appropriately defined Riemannian manifold. We provide numerical simulations which allow us to study and characterize the effect of the mirror potential on the performance of networks trained with SMD for some binary classification problems. 
\end{abstract}


\section{Introduction}\label{sec:intro}
Machine learning models have been successfully used in a wide variety of applications ranging from spam detection, to image classification, to natural language processing. Despite this great success, our theoretical understanding on why various machine learning methods demonstrate the performances they do is still at an early stage.

To obtain some understanding of the behavior of machine learning algorithms we shall focus on the structure/architecture of the model, as well as the training mechanism that generates a suitable model with respect to the available data. With regards to the architecture, a common theme among most modern machine learning models (e.g. artificial/convolutional/recurrent neural networks) is that they repetitively use simple blocks such as perceptrons in ANNs, and  convolution filters in CNNs. The idea of combining  multiple simple models has its roots in classical statistics and is known as boosting. The justification is that a simple  model often possesses a low variance while its bias can be potentially high; therefore, combining multiple such models can reduce the bias while keeping the variance low. 

The optimization algorithm maps the training data to model parameters. Its role is especially critical in modern setups where the number of parameters can be overwhelmingly larger than the number of inputs. In such overparameterized settings,  there are often many models that perfectly fit the training data. Thus, the specific choice of the optimization algorithm determines the resulting model in the interpolating regime. In fact, understanding the connection between the optimization algorithm and the generalization performance of the resulting model is an important open problem.

In this paper we attempt to study these aspects by analyzing the convergence behavior of the stochastic mirror descent (SMD) iterates on a general class of models, known as \emph{average ensemble models} where the output is generated by taking an average of similar models that only differ in their choice of parameters. Two-layer neural networks, e.g., can be viewed as a special case of the average ensemble.

In particular, we address the impact of the potential function used in SMD on the parameters of the average ensemble model. Introduced by \cite{nemirovskii_problem_1983} for convex optimization problems with underlying geometric structure, SMD is as a generalization of stochastic gradient descent (SGD) where the updates are \emph{mirrored} to a dual domain using a \emph{potential function}. It has been shown that choice of the potential function plays a pivotal role in determining the implicit bias of the SMD in the interpolating regime \cite{azizan_stochastic_2019-1,gunasekar_characterizing_2020}. 

\subsection{Contributions and Prior Work}

In this paper, we develop a mean-field type result for average ensemble models trained with SMD. We consider a general class of convex loss functions that includes the squared loss as a special case. In particular, we derive a partial differential equation that describes the "time evolution" of the distribution of the parameters in the network as training progresses. This distribution can be used to compute the risk, or generalization error, of the trained model. Our result generalizes similar mean-field results for SGD, with the difference being that the Hessian of the potential function appears. We further give a geometric interpretation at the "distributional" level by showing an equivalent formulation of our result as a geometric flow on an appropriate Riemannian manifold. We finally provide numerical results to illustrate the applicability of the method and to showcase the effect of the potential function on the performance and implicit bias of the resulting trained network.

A mean-field characterization of two-layer neural networks trained with SGD under quadratic loss was studied by \cite{mei_mean_2018}, \cite{sirignano_mean_2020}, and \cite{rotskoff2019trainability}. A similar line of work is given by \cite{chizat_global_2018} where more general loss functions are considered for gradient descent algorithm. 
Mean field results with similar flavor to ours have been given in \cite{raginsky_continuous-time_2012} and \cite{borovykh_stochastic_2020} in the context of distributed convex optimization. These papers consider the problem of minimizing a fixed objective function with noisy information by many "interacting optimizers" coupled through an interaction matrix. Despite being insightful, the setting presented in these papers does not  fit into the case of learning with average ensemble models.
An important connection between mirror descent and natural gradient descent on dual (Riemannian) space was established by \cite{raskutti_information_2014}. Works by \cite{gunasekar_characterizing_2020} and \cite{gunasekar_mirrorless_2020} advanced this geometric viewpoint by showing a correspondence between continuous-time mirror descent and Riemannian gradient flow on the primal domain governed by the Hessian matrix of the potential function.

\section{Preliminaries}\label{sec:prelim}
\subsection{Notations}
\label{subsec:notations}
We gather here the basic notations used. Vectors are presented with bold lower-case letters, and bold upper letters are reserved for matrices. For a vector $\v$, $v_i$ denotes its $i^{\text{th}}$ entry, and $\| \v \|_p$ is  its $\ell_p$ norm. $\mathbb S^d_{++}$ is the set of symmetric and positive-definite matrices with dimension $d$.

For a set $\S$ in $\R^d$ and $k \in \N$, $\C_b^{k}(\S)$ denotes the space of bounded and $k$-times differentiable functions with continuous $k^{th}$ derivative, $\PP(\S)$ denotes the space of Borel probability measures on $\S$ and $\PP_m(\S)$ denotes probability measures with bounded $m^{\text{th}}$ moment.

For a random vector $\mathbf Z$ in $\R^p$, $\mu(d\mathbf z)$ indicates its probability measure. $\delta_{\x}$ is a Dirac mass at point $\x$. Integral of a function $f:\S\rightarrow \R$ with respect to a measure $\rho$ on $\S$ is denoted as,
\begin{equation}
    \langle \rho, f \rangle :=  \int f(\x) \rho (d\x).
\end{equation}
Let $\{\rho_n\}_{n\in \N}$ be a sequence of probability measures defined on $\S\subseteq \R^d$. The sequence is said to be {\textit{weakly converging}} to the measure $\rho^\star$ if and only if, $\underset{n\rightarrow \infty}\lim  |\langle \rho_n, f \rangle - \langle \rho^\star, f \rangle| = 0$,
for all $f \in \C_b(\S)$.
\subsection{Problem Setup}
\label{subsec:setup}
Consider an online learning setting where we sequentially observe a fresh sample of data, i.e., $(\mathbf x^{k}, y^{k})$ for $k=1,2,\ldots$. Here $\x^k \in \mathcal X \subseteq \R^p$ is the feature vector, and $y^k \in \mathcal Y \subseteq \R$ is the label. As a shorthand, we will interchangeably use $\mathbf z:=(\mathbf x, y)$. Our goal is to learn a model that generates an estimate of the label given the feature vector as its input. A common approach for modeling is known as the average ensemble, where the output is generated by taking the average of simpler models that only differ in their parameters. This approach is advantageous from the bias-variance perspective. Since simpler models have a low variance (yet potentially a high bias), the average ensemble  generates a low-variance estimate of the labels with a reasonably good bias.\\
Given a parameter space $\S \subset \R^d$ and $\sigma:\X \times \S \rightarrow \Y$, consider the following parametric family of model functions,
\begin{equation}
    \mathcal H_{\sigma} = \{\sigma(\cdot, \bm\theta):\X \rightarrow \Y;\; \bm \theta \in \S \}.
\end{equation}
This parametric representation of model functions has applications in various settings. For instance, by setting $d=p$ and defining $\sigma(\v_1, \v_2) = \sigma^\star(\v_1^T\v_2)$, the class $\mathcal H_{\sigma}$ represents the generalized linear models with nonlinearity $\sigma^\star:\R\rightarrow\R$.

The average ensemble model generates an estimate by taking the average of $n$ models in $\mathcal H_{\sigma}$. In other words, for $\x \in \X$ the estimate is computed as follows,
\begin{equation}
    \yh = h_n(\x;\, \bm\Theta) := \frac{1}{n}\sum_{i=1}^{n} \sigma (\x, {\bm \theta}_i),
\end{equation}
where $\bm \Theta = \begin{pmatrix} \bm \theta_1,~\bm \theta_2,\ldots, ~\bm \theta_n\end{pmatrix} \in \S^n$.\\ \\
As explained above, average ensemble models are used in many practical settings. A popular instance of such models is the two-layer neural network, where ${\bm \theta}_i = \begin{bmatrix}\bar{\bm \theta}_i,~ w_i \end{bmatrix}^T$, and the output is computed as,
\begin{equation}
    \yh_{NN} = \sum_{i=1}^{n} \frac{w_i}{n}\bar \sigma (\x, \bar {\bm \theta}_i),
\end{equation}
where $\mathbf w \in \R^n$ represents the weights of the last layer of the network and $\bar \sigma (\cdot)$ is the activation function.

Given a loss function $\ell:\R^2\rightarrow \R$ and a training dataset $\{(\x^k,y^k)\}_{k=1}^K$, our goal is to minimize the empirical risk
\begin{equation}
    \inf_{\Theta \in \S^n}\left\{ \hat{R}_n(\Theta) := \frac{1}{K} \sum_{k=1}^{K} \ell(y^k,\, h_n(\x^k;\, \Theta)) \right\}~,
\end{equation}
over parameters $\bm \Theta$. We assume that the training data points are generated independently from a distribution, i.e., $(\x^k, y^k) \sim \mu(d\mathbf z)$. The loss function $\ell:\R^2\rightarrow \R$ is assumed to be convex w.r.t. its second argument. Important examples are the quadratic loss, $\ell(u,v) = \frac{1}{2}(u-v)^2$, and the logistic loss, $\ell(u,v) = \log\big(1+\exp(-u\times v)\big)$. However, the problem is non-convex whenever $\sigma$ is. 
One can rewrite the average ensemble model as an integral over the empirical distribution of the parameters,  $\hat \rho_n(d\bm\theta):= \sum_{i=1}^n\frac{1}{n}\delta_{\bm\theta_i}(d\bm\theta)$, i.e., 
\begin{equation}\label{eq:integral_form_1}
    h_n(\x;\, \bm\Theta) = \langle\hat \rho_n, \sigma(\x, .)\rangle.
\end{equation}
This new form leads us to define ensemble averages by any probability distribution on $\S$. Namely, we define
\begin{equation}\label{eq:integral_form_2}
    h(\x;\, \rho) = \langle \rho, \sigma(\x, .)\rangle,
\end{equation}
for any $\rho \in \PP(\S)$. We call this the \textit{mean-field ensemble model} and the former the  \textit{finite ensemble model}.  It is easy to check that $h_n(\x;\, \bm\Theta) = h(\x;\, \hat \rho_n)$. Note that whenever the loss function $\ell$ is convex, the mean-field ensemble model function $h(\x;\, \cdot):\PP(\S) \rightarrow \Y$ is also convex in the space of probability measures $\PP(\S)$ regardless of $\sigma(\cdot)$.

\subsubsection{Stochastic Mirror Descent Updates}
Mirror descent algorithms introduced by ~\cite{nemirovskii_problem_1983}, and their stochastic variants, are commonly used iterative methods that exploit a potential function to impose certain attributes in the resulting model. Assume $\mathcal X = \R^p$, $\mathcal Y = \R$, and $\S = \R^d$. Here, we consider the mirror descent updates initialized at $\bm \Theta^{0} \in \S^n$ where, for $i=1,2,\ldots,n$, the updates are defined as,
\begin{equation}
    \label{eq:SMD_updates_1}
    \nabla \psi(\bm\theta_i^{k+1}) = \nabla \psi(\bm\theta_i^{k}) -  \frac{\tau}{n} \nabla_{\bm\theta_i} \ell(y^{k+1},\, h_n(\x^{k+1};\, \Theta^k)),
\end{equation}
and $k$ runs from $0$ to $K-1$. Here $\psi(\cdot)$ is a strongly convex and differentiable potential function, $\tau$ a fixed scaling of the step-size, and $\bm{\Theta}^k$ the parameters after the $k^{\text{th}}$ update. Strong convexity of $\psi(\cdot)$ ensures the invertibility of its gradient which makes the updates well-defined (see~\cite{bubeck_convex_2015}).

After evaluating the gradient of the loss function (w.r.t. $\bm\theta_i$) and using the identity~\eqref{eq:integral_form_1}, we can rewrite the SMD updates for $i=1,2,\ldots,n$ and $k=0,\dots,K-1$ as
\begin{equation} \label{eq:SMD_updates_2}
    \begin{cases}
    \hat\rho^k_n = \frac{1}{n}\sum_{i=1}^n\delta_{\bm\theta_i^k}~,\\
    \nabla \psi(\bm\theta_i^{k+1}) = \nabla \psi(\bm\theta_i^{k}) +  \frac{\tau}{n} \mathbf F(\bm \theta_i^k, \, \hat{\rho}^k,\, \mathbf z^{k+1})~,
    \end{cases}
\end{equation}
where ${\hat \rho}^k$ denotes the empirical distribution of the parameters after the $k^{\text{th}}$ update. The function $\mathbf F$ is the gradient of the loss w.r.t. $\bm\theta$, which is  defined as,
\begin{equation} \label{eq:gradient_loss}
    \mathbf F(\bm\theta, {\rho}, \z) :=  -\partial_2 \ell\big(y; \langle \rho,~ \sigma(\x, \cdot)\rangle \big) \nabla_{\bm\theta} \sigma(\x,\bm\theta),
\end{equation}
where $\partial_2$ indicates the derivative w.r.t. the second argument. It is worth noting that the potential function, $\psi(\cdot)$, is often chosen to enforce some structure on the resulting parameter. Setting  $\psi(\cdot)=\frac{1}{2}\|\cdot\|_2^2$ gives the SGD iterates.

\emph{Our goal is to compute the (converging) distribution of the parameters in the asymptotic regime where $n, K$ go to infinity at a fixed ratio.} This distribution plays a pivotal role in understanding the generalization behavior of the resulting model. In particular, the expected risk can be computed as,
\begin{equation}
    \label{eq:standard_risk}
    R(\rho) := \mathbb E_{(\x, y) \sim \mu(d\mathbf z)} \big[\ell \big(y; \langle \rho,~ \sigma(\x, \cdot)\rangle\big)\big]~,
\end{equation}
where $\rho:=\rho(d\bm \theta)$ denotes the (empirical) distribution.

\section{Main Result}\label{sec:result}
In this section, we present the main result of the paper, i.e., the characterization of the converging distribution of the SMD iterates. The resulting distribution is given as the solution to a continuous-time nonlinear PDE. We consider the converging distribution in the asymptotic regime where $K,n\rightarrow \infty$ at a fixed ratio, $\delta:=\frac{K}{n}\in\R_{++}$.

In Section~\ref{subsec:discrete_to_continous}, we explain how to map SMD updates to a continuous-time process on probability distributions in $\R^d$. Consequently, a PDE will be derived in Section~\ref{subsec:differential_equation} whose evolution is captured by the Hessian of the potential function, and the function $\mathbf F$ defined in~\eqref{eq:gradient_loss}. Finally, Section~\ref{subsec:main_result} incorporates the main result of the paper which indicates that the distribution of the mirror descent updates weakly converges to the solution of the PDE.

We require the following set of assumptions.
\begin{assump} \label{thm:assump}
\mbox{}
\begin{itemize}
\item[(i)] $\sigma:\R^p\times\R^d \rightarrow \R$ is bounded and $\sigma(\x,\,\cdot) \in \C_b^2(\R^d)$ for all $\x \in \R^p$.
\item[(ii)] The loss function $\ell:\R\times\R \rightarrow \R$ is convex and $\C^2(\R)$ w.r.t. its second argument.
\item[(iii)]  The mirror function $\psi:\R^d \rightarrow \R$ is $\lambda-$strongly convex and $\psi \in \C^2(\R^d)$.
\item[(iv)] The initial parameters $\bm \Theta^0 = \begin{pmatrix} \bm \theta_1^0,~\ldots, ~\bm \theta_n^o\end{pmatrix}$ are sampled i.i.d. from a distribution $\rho^0 \in \PP_4(\R^d)$.
\item[(v)] The data distribution has finite $4^{\text{th}}$-order moments, i.e., $\mu \in \PP_4(\R^{p+1})$
\end{itemize}
\end{assump}
Assumptions $(i)$ and $(v)$ are required for well-definiteness of expectations and together with $(iv)$, they will guarantee our convergence result. Convexity of loss function $\ell$ ensures convexity of the expected risk of the mean-field ensemble which model \eqref{eq:standard_risk}. Assumptions $(ii)$ and $(iv)$ guarantee uniqueness of the resulting PDE. Finally, assumption $(iii)$ is needed for non-singularity of the Hessian of the mirror function.

\subsection{Continuous Time Viewpoint}\label{subsec:discrete_to_continous}

Recall from the previous section ${\hat \rho}^k_n$, for $k=0,1,\ldots, K-1$,  is defined as the empirical measure of the parameters after $k^{\text{th}}$ mirror update, i.e, $
    {\hat \rho}^k_n(d\bm\theta) :=  \frac{1}{n}\sum_{i=1}^n\delta_{\bm\theta_i^k}(d\bm\theta)$.

To characterize the converging distribution of SMD, we exploit a similar approach as~\cite{wang2017scaling} and~\cite{sirignano_mean_2020} by viewing $\{\hat\rho^k_n\}_{k=0}^{K-1}$ as a \emph{Markov process in the state space} $\mathcal P(\R^d)$. The Markov property follows from~\eqref{eq:SMD_updates_2}, where $\{\bm\Theta^{k}\}_{k=0}^{K}$ forms an exchangeable Markov process, i.e., any permutation of the indices $\begin{pmatrix} \bm \theta_1^k,~\ldots, ~\bm \theta_n^k\end{pmatrix}$ will leave their joint distribution intact. \\ \\
We introduce $\bar\rho_n(t)$ which is a continuous-time embedding of the discrete-time process $\{\hat \rho^k_n\}_{k=0}^{K-1}$, defined as,
\begin{equation} \label{eq:CT_embedding}
    \bar\rho_n(t) := \hat \rho_n^{\lfloor{nt/\tau}\rfloor}~,\qquad t\in [0, T],
\end{equation}
where $T:=K\times \frac{\tau}{n}=\delta\tau$ is a positive constant. $\bar \rho_n(t)$ is a piecewise constant process that is right-continuous with left limits (RCLL). This process can be viewed as an element in the Skorokhod space $\mathcal D([0, T]; \mathcal P(\S))$\footnote{For a set $\X$ and $T>0$,the Skorokhod space $\mathcal D([0,T], \X)$ is the space of \textit{càdlàg} functions from the interval $[0,T]$ to $\X$ which are right-continuous with left-limits.}.  
We will show in Section~\ref{subsec:main_result} that $\{\bar \rho_n(t)\}_{n\in \N}$ has a limit as $n\rightarrow \infty$.

\subsection{A Partial Differential Equation}\label{subsec:differential_equation}
Here we introduce a non-linear partial differential equation (PDE) whose solution will determine the converging distribution of SMD updates. Let  $\mathbf H:\R^d\rightarrow \mathbb S^d_{++}$ be the Hessian operator of the mirror potential function $\psi(\cdot)$, i.e.,
\begin{equation}
    \mathbf H(\bm \theta) := \frac{\partial^2}{\partial {\bm\theta}^2}\psi(\bm \theta)~~,~~\bm\theta\in\R^d.
\end{equation}
Note that $\mathbf H(\bm\theta) \succeq \lambda \mathbf I_d$ due to the strong convexity of the mirror potential. Consider the following PDE on probability measures with the initial condition $\rho_0 \in \mathcal P(\R^d)$,
\begin{equation}\label{eq:PDE}
    \partial_t \rho_t = -\nabla_{\bm\theta} \cdot \left(\rho_t \mathbf H(\bm \theta)^{-1} \mathbf v(\bm\theta, \rho_t)\right),
\end{equation}
where $\nabla_{\bm\theta} \cdot\mathbf J$ denotes the divergence of the vector field $\mathbf J$\footnote{For a vector field $\mathbf F:\R^d\rightarrow \R^d$, defined as $\mathbf F(\x)=\begin{bmatrix}F_1(\x),F_2(\x),\ldots,F_d(\x)\end{bmatrix}^T$, the divergence is defined as, $$\nabla_{\x}\cdot \mathbf F  = \frac{\partial F_1}{\partial {x_1}} + \frac{\partial F_2}{\partial {x_2}}+\ldots + \frac{\partial F_d}{\partial {x_d}}.$$}, and $\mathbf v:\R^d\times \mathcal P(\R^d)\rightarrow \R^d$ is defined as,
\begin{equation}
    \begin{aligned}
    &&\mathbf v(\bm \theta, \rho)&:=\mathbb E_{(\x, y) \sim \mu(d\mathbf z)}\left[\mathbf F(\bm\theta, {\rho}, \x, y)\right],\\
    &&&~=\int \mathbf F(\bm\theta, {\rho}, \x, y) \mu(d\mathbf z)~,
    \end{aligned}
    \label{eq:velocity_definition}
\end{equation}
where $\mathbf F$ is the negative gradient of the loss function defined in~\eqref{eq:gradient_loss}. An alternative but equivalent definition for the velocity field $\v$ is given as
\begin{equation}\label{eq:velocity}
    \mathbf v(\bm\theta, \rho) = -\nabla_{\bm\theta} \frac{\delta R}{\delta \rho}(\bm\theta, \rho),
\end{equation}
where $\frac{\delta R}{\delta \rho}$ denotes the functional (Fréchet) derivative of the expected risk functional $R(\rho)$~\eqref{eq:standard_risk}. 

We note that the PDE~\eqref{eq:PDE} should be understood in the weak sense, i.e., $\rho_t$ is a solution of~\eqref{eq:PDE} if it solves the following equation for any test function $f \in \C_b^2(\S)$
\begin{equation}\label{eq:weak_PDE}
    \begin{aligned}
    \frac{d}{d\,t}\langle \rho_t, f \rangle &= \langle \rho_t, \nabla f(\cdot)^T \mathbf H(\cdot)^{-1} \mathbf v(\cdot, \rho_t) \rangle,\\
    &=\int \nabla f(\bm\theta)^T \mathbf H(\bm \theta)^{-1} \mathbf v(\bm\theta, \rho_t) \rho_t (d\bm\theta).
    \end{aligned}
\end{equation}
The PDE~\eqref{eq:PDE} can be interpreted as a continuity equation describing the flow of probability distribution driven by the "distorted" drift/velocity field $\mathbf H^{-1}(\bm\theta)\mathbf v(\bm\theta, \rho)$. The distortion of the drift term is due to inverse Hessian of the mirror potential while the impact of the data distribution is captured by $\mathbf v$. In Section~\ref{sec:Riemannian}, we provide an equivalent formulation for~\eqref{eq:PDE} in which the impact of mirror is totally captured by a newly defined Riemannian metric. This leads to a coordinate-free and abstract reformulation of~\eqref{eq:PDE}.
\begin{remark} \label{rem:PDE}
The PDE in~\eqref{eq:PDE} is presented in a general form. For specific choices of the loss function, the drift term $\mathbf v$ can be simplified. There are variety of approaches to solve this nonlinear PDE; a common one is fixed-point iteration starting from a (reasonable) guess for the solution in the path space and making iterative updates according to~\eqref{eq:PDE}. Discussion of methods for solving PDEs is beyond the scope of this paper, and we refer the reader to~\cite{sznitman_topics_1991} for more details.
\end{remark}
\subsubsection{Example: Squared loss}\label{ssubsec:square_loss}
The PDE introduced in~\eqref{eq:PDE} applies to any convex loss function. The most popular choice in regression problems is the squared loss. Here, we present a simplified version of the velocity field $\v$ for the squared loss. To this end, define the following two auxiliary functions.
\begin{equation}
    \begin{cases}
    V(\bm\theta) = -\mathbb E_{(\x, y) \sim \mu(d\mathbf z)}\left[y\cdot\sigma(\x, \bm\theta)\right]~,\\
    U(\bm\theta, \bm \theta') = \mathbb E_{(\x, y) \sim \mu(d\mathbf z)}\left[\sigma(\x, \bm\theta)\cdot \sigma(\x, \bm\theta')\right]~.
    \end{cases}
\end{equation}
Then velocity field $\mathbf v(\bm \theta, \rho)$ can be written as,
\begin{equation}
    \label{eq:squared_loss}
    \mathbf v(\bm \theta, \rho)= -\nabla_{\bm\theta}V(\bm\theta) - \langle\rho,\nabla_{\bm\theta}U(\bm\theta, \cdot)\rangle. 
\end{equation}
One main advantage of having this equivalent formulation is that the auxiliary functions can be computed for a broad class of distributions $\nu$, and nonlinear functions $\sigma$. For our numerical simulations in Section~\ref{sec:simul} we compute $U$ and $V$ for  $\sigma(\x,\bm\theta) = \text{erf}\left(\x^T{\bm\theta}\right).$\footnote{For $z\in\R$, $\text{erf}(z)$ is defined as, $$\text{erf}(z):= \frac{2}{\sqrt{\pi}} \int_{0}^{z} e^{-x^2}dx.$$} Analytical expressions for $V$ and $U$ for single-hidden-layer networks, Gaussian mixture data models, and general $\sigma$ can be derived, though we shall not include them here for lack of space. 

\subsection{Converging distribution}\label{subsec:main_result}

 Our main result is Theorem~\ref{thm:main}, which states that the time evolution of the empirical distribution of the parameters converges to the solution of the PDE~\eqref{eq:PDE} (initialized with ${ \rho}^0$). Equivalently, the empirical distribution after $k = r\cdot n$ updates can be approximated the solution of the PDE at time $t = r\cdot \tau$.

\begin{theorem}[Converging Distribution of SMD]
\label{thm:main}
Let $\{(\x^k, y^k)\}_{k=0}^{K-1}$ denote a dataset where $(\x^k, y^k)\sim \mu (d\mathbf z)$. Consider the stochastic mirror descent updates in~\eqref{eq:SMD_updates_2} and assume that Assumption~\ref{thm:assump} holds. As $n,K\rightarrow \infty$ at a fixed ratio $\delta=\frac{K}{n}$, the continuous-time process $\{\bar \rho_n(t):t\in[0, T]\}$ defined in~\eqref{eq:CT_embedding} converges weakly to $\{\rho_t^{\star}:t\in[0, T]\}$ in the Skorokhod path space $\mathcal D([0, T]; \mathcal P(\S))$ where $\rho_t^\star$ is the solution of the following PDE for $t\in [0,T]$,
\begin{equation*} 
    \partial_t {\rho_t^\star} = -\nabla_{\bm\theta} \cdot \left({\rho_t^\star} \mathbf H(\bm \theta)^{-1} \mathbf v(\bm\theta, {\rho_t^\star})\right),
\end{equation*}
with $\rho_0^\star = {\hat\rho}^0$. In other words, the empirical distribution of parameters resulting from SMD iterates converges to the solution of the partial differential equation~\eqref{eq:PDE}.
\end{theorem}

\begin{remark}\label{rem:convergence}
As a side technical note, the convergence in Theorem~\ref{thm:main} is in Skorokhod distance, which is a distance defined on \textit{càdlàg} processes. For a given  $f\in \C_b^2(\R^d)$, denote by $\bar\rho_n[f] := \{\langle \bar{\rho}_{n}(t),f\rangle \, :\, t \in [0,T]\}$ and $\rho^\star[f] := \{\langle {\rho_t^\star},f\rangle \, :\, t \in [0,T]\}$ the processes generated by projecting function $f$ onto $\mathcal D([0, T]; \mathcal \R)$. Theorem~\ref{thm:main} states that the (Skorokhod) distance between these two  \textit{càdlàg} processes converges to zero in probability as $n\rightarrow \infty$, i.e.,
\begin{equation}
    d_{\mathcal D[0,T]} \left(\bar\rho_n[f]
    , \,  \rho^\star[f] \right) \overset{\mathbb  P}\longrightarrow ~0~,
\end{equation}
where $d_{\mathcal D[0,T]}$ is the Skorokhod distance on $\mathcal D([0, T]; \mathcal \R)$. We refer the interested reader to ~\cite{billingsley_convergence_1999} and \cite{kallenberg_foundations_2010} for a thorough discussion on convergence of probability measures and random processes.
\end{remark}
An outline of the proof is provided in Section \ref{sec:proof}. \\
Notice that when the mirror potential is chosen as $\psi(\cdot)=\frac{1}{2}\|\cdot\|_2^2$, its Hessian becomes $\mathbf H(\bm\theta) = \mathbf I_d$, the identity matrix and the PDE \eqref{eq:PDE} takes the following simpler form.
 \begin{equation}\label{eq:SGD_PDE}
    \partial_t \rho_t = -\nabla_{\bm\theta} \cdot \left(\rho_t  \mathbf v(\bm\theta, \rho_t)\right).
 \end{equation}
This corresponds to the specific case of stochastic gradient descent (SGD). A neat interpretation of \eqref{eq:SGD_PDE} as a Wasserstein gradient flow\footnote{For a given a metric space $(\X,d)$, the Wasserstein-2 distance, $W_2:\PP(\X)\times \PP(\X) \rightarrow \R$ on $\PP(\X)$ is defined as 
\begin{equation*}
W_2(\rho_1, \rho_2) = \left(\inf_{\gamma \in \Gamma(\rho_1,\rho_2)} \int d(x_1,x_2)^2 \gamma(dx_1,dx_2)\right)^{1/2}
\end{equation*}
where $\Gamma(\rho_1,\rho_2)$ is the set of probability distributions on $\PP(\X\times\X)$ with marginals $\rho_1$ and $\rho_2$.}  
on the space of probability distributions $\PP(\R^d)$ is given by \cite{mei_mean_2018}, \cite{sirignano_mean_2020}, \cite{chizat_global_2018}, and \cite{rotskoff2019trainability}. 

At a high level picture, this means that the solution of \eqref{eq:SGD_PDE}, $t\mapsto \rho_t$ follows a trajectory on $\PP(\R^d)$ that minimizes the expected risk functional $R(\rho)$ as time passes. This view also gives a clear meaning to velocity field $\v(\bm\theta, \rho)$. Recalling the alternative definition given in \eqref{eq:velocity}, we can view $\mathbf v(\bm\theta, \rho) = -\nabla_{\bm\theta} \frac{\delta R}{\delta \rho}(\bm\theta, \rho)$ as the direction of "steepest descent" for the expected risk functional $R(\rho)$ in the Euclidean sense. We refer the reader to \cite{ambrosio_gradient_2008} for a comprehensive study on gradient flows. 

When a general mirror potential is considered, the inverse Hessian $\mathbf H^{-1}$ in the main equation $\eqref{eq:PDE}$ is posing a challenge to this view since  $\mathbf H^{-1}(\bm\theta)\mathbf v(\bm\theta, \rho)$ is not the steepest descent direction in the Euclidean way. In the next section, we introduce an appropriately defined Riemannian manifold in order to obtain a similar steepest descent interpretation. 

\section{Riemannian Formulation}\label{sec:Riemannian}
In this section, we give an informal interpretation of equation \eqref{eq:PDE} in a geometric sense. The main idea is to reformulate PDE \eqref{eq:PDE} on a Riemannian manifold on which the metric tensor is defined by $\mathbf H(\bm\theta)$.  

\begin{theorem}
Assume the entries of $[\mathbf H(\bm\theta)]_{ij}:\R^d \rightarrow \R$ be smooth functions of $\bm\theta$, and equip the parameter space $\S=\R^d$ with a Riemannian metric $g_{ij}(\bm\theta) := [\mathbf H(\bm\theta)]_{ij}$. Assume, in addition, that the initial distribution admits a density with respect to volume measure on $(\S, g)$ as  $\rho_0(\bm\theta)$ = $p(\bm\theta, 0) dV_g(\bm\theta)$. Then, PDE \eqref{eq:PDE} takes the following form on the Riemannian manifold $(\S, g)$
\begin{equation} \label{eq:PDE_riemann}
    \frac{\partial p(\bm\theta, t)}{\partial t} = \divv_{g}\left(p(\bm\theta, t) \grad_g\frac{\delta R}{\delta \rho}(\bm\theta, p_t) \right).
\end{equation}
where $\divv_g$ and $\grad_g$ denote the divergence and gradient operators on $(\S,g)$, respectively.
\end{theorem}

The insight behind this theorem is to define a new geometry on $\S=\R^d$ from the Hessian $\mathbf H(\bm\theta)$ of the potential. Defining Riemannian metrics from Hessians of smooth functions is a widely-used technique (\cite{shima_hessian_1997}). 

For a comparison, we rewrite the PDE \eqref{eq:SGD_PDE} by inserting the definition of $\v$ as
 \begin{equation}\label{eq:SGD_PDE_2}
    \partial_t \rho_t = \nabla_{\bm\theta} \cdot \left(\rho_t  \nabla_{\bm\theta} \frac{\delta R}{\delta \rho}(\bm\theta, \rho_t)\right).
 \end{equation}
When contrasted with the above equation derived for Euclidean mirror potential, the PDE \eqref{eq:PDE_riemann} has the same form where divergence and gradient operators on Euclidean space are replaced by their Riemannian counterparts. 

In lieu of this observation, we give an interpretation of \eqref{eq:PDE_riemann} as a geometric flow describing the time evolution of parameter distribution on the Riemannian manifold defined by the Hessian of the potential. In this new geometry, steepest descent directions no longer follow straight lines, but rather follow geodesics\footnote{Geodesics are the shortest distance paths connecting points on a Riemannian manifold.} imposed by the mirror potential.
\begin{proof}
We start by defining the gradient operator on $(\S, g)$. For a smooth function $f\in \C^{\infty}(\S)$, its gradient at a point $\bm\theta \in \S$ with respect to metric $g$ is defined as
\begin{equation}
    \grad_g f :=  \mathbf{H}(\bm\theta)^{-1} \nabla_{\bm\theta}f (\bm\theta)
\end{equation}
We thus obtain the following identity for  $\mathbf{H}^{-1}\mathbf v $ 
\begin{equation}
    \mathbf{H}(\bm\theta)^{-1}\mathbf v(\bm\theta, \rho) = -\grad_g  \frac{\delta R}{\delta \rho}(\bm\theta, \rho).
\end{equation}
Inserting this back to equation \eqref{eq:PDE} gives
\begin{equation} \label{eq:PDE_riemann_v2}
    \partial_t \rho_t = \nabla_{\bm\theta} \cdot \left(\rho_t  \grad_g \frac{\delta R}{\delta \rho}(\bm\theta, \rho_t)\right).
\end{equation}
 For a smooth vector field $\mathbf{J}(\bm\theta)$, the Riemannian divergence operator on $(\S, g)$ is defined as
\begin{equation}
    \divv_g \mathbf J :=  \frac{1}{\sqrt{\dett{\mathbf{H}(\bm\theta)}}} \nabla_{\bm\theta} \cdot \left(\sqrt{\dett{\mathbf{H}(\bm\theta)}} \mathbf{J}(\bm\theta) \right)
\end{equation}
Note that $\sqrt{\dett{\mathbf{H}(\bm\theta)}}$ also appears in  the definition of the volume measure $dV_g(\bm\theta)$ on $(\S,g)$ as,
\begin{equation}
    dV_g(\theta) = \sqrt{\dett{\mathbf{H}(\bm\theta)}} d\bm\theta.
\end{equation}
In other words, infinitesimal volumes on $(\S, g)$
are distorted by a factor of $\sqrt{\dett{\mathbf{H}(\bm\theta)}}$. 
Inserting the new divergence operator on $(\S,g)$ to \eqref{eq:PDE_riemann_v2}, we rewrite it as
\begin{equation} \label{eq:PDE_riemann_v3}
    \partial_t \frac{\rho_t}{\sqrt{\dett{\mathbf{H}(\bm\theta)}}} =  \divv_g \left(\frac{\rho_t}{\sqrt{\dett{\mathbf{H}(\bm\theta)}}}  \grad_g \frac{\delta R}{\delta \rho}(\bm\theta, \rho_t)\right).
\end{equation}
Using the assumption on the initial distribution, the solution of \label{eq:PDE_riemann_v3} at any time step can be expressed similarly 
\begin{equation}
    \begin{aligned}
        \rho_t(d\bm\theta) &= p(\bm\theta,t)dV_g(\bm\theta), \\
        &= p(\bm\theta,t)\sqrt{\dett{\mathbf{H}(\bm\theta)}}d\bm\theta.
    \end{aligned}
\end{equation}
Then, the resulting \eqref{eq:PDE_riemann} follows immediately. 
\end{proof}

\section{Numerical Simulations}\label{sec:simul}
In order to illustrate the applicability of our theoretical results, we have performed several simulations by numerically solving the PDE \eqref{eq:PDE}. Due to lack of space, we only include a single illustrative example here.

\begin{figure}
    \centering
    \includegraphics[width =0.8\linewidth]{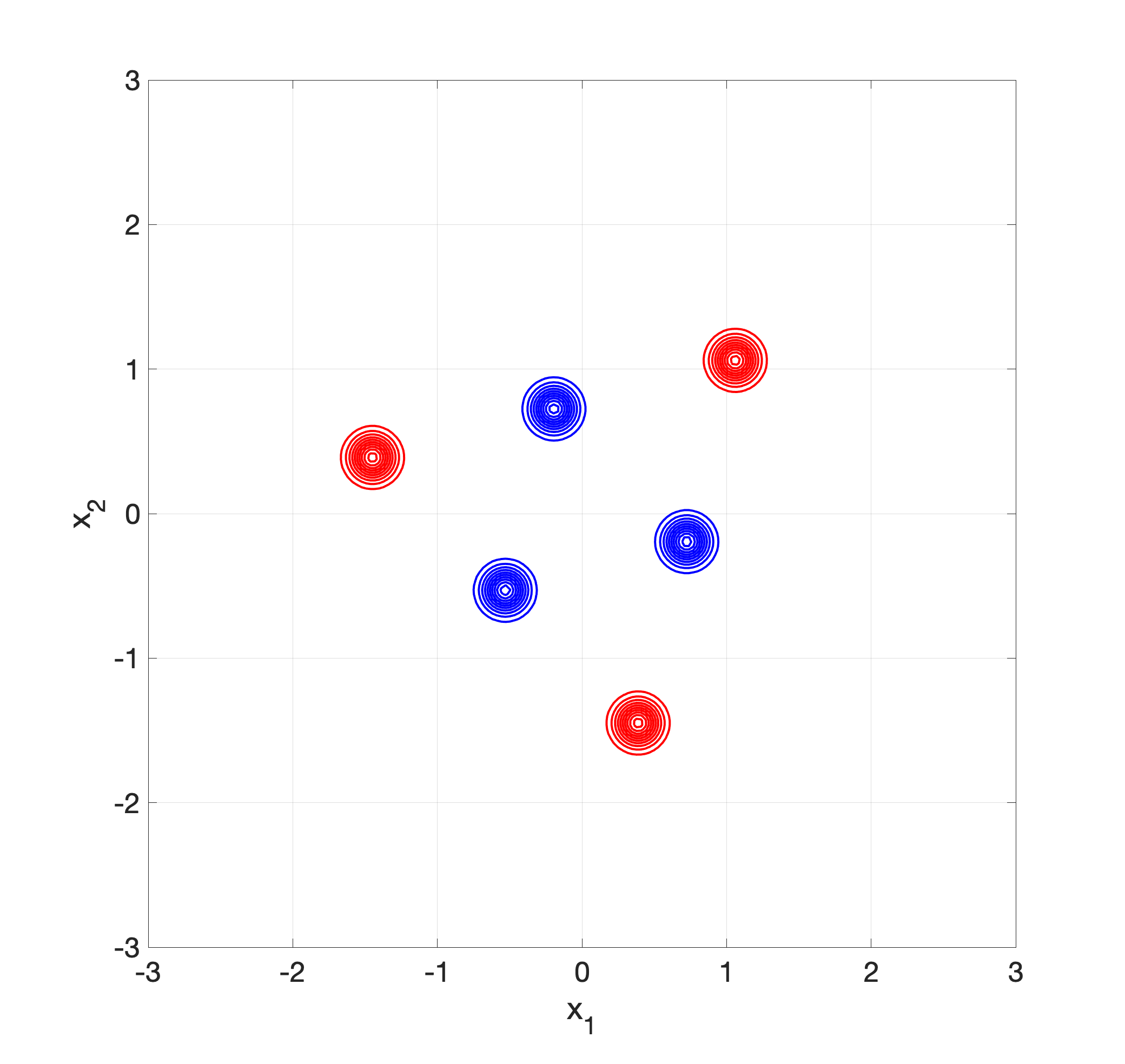}
    \caption{The centers of the Gaussian mixture model for class 1 (red) and class 2 (blue). Note that the three centers for each class form an equilateral triangle with geometrical center at the origin. The centers for class 2 are the midpoint of the edges connecting the centers of class 1. Finally, the centers of the two classes can be separated by three lines with relative angles $120^o$ passing through the origin.  }\label{fig:data}
\end{figure}

Consider a binary classification problem with two-dimensional inputs and a data distribution given by
\begin{equation}
    \begin{cases} 
      y = +1,\; \x \sim \text{GM}_1,\quad \text{with probability $\frac{1}{2}$},  \\
      y = -1,\; \x \sim \text{GM}_2,\quad \text{with probability $\frac{1}{2}$}
   \end{cases}
\end{equation}
where $\text{GM}_1$ and $\text{GM}_2$ are two different Gaussian mixtures, with three centers each, and isotropic covariance matrices
\begin{equation}
    \text{GM}_i(\x) = \frac{1}{3}\sum_{j=1}^{3} \normal(\x \,|\,\mathbf m_i^j,\, s^2\mathbf I_2 ), \; \; i=1,2
\end{equation}
Here $s=0.1$.  The exact position of the centers $\{\mathbf m_i^1, \mathbf m_i^2, \mathbf m_i^3\}$ are shown in Figure~\ref{fig:data} and are chosen so that they form two equilateral triangles (one for each class) with geometric centers at the origin. Most importantly, centers of the second class (blue) are exactly the midpoints of the edges of the triangle formed by the centers of the first class (red). Due to the symmetry in the data distribution, it is possible to separate the two pairs of centers for each class using three lines through the origin. 

We set the function $\sigma(\x, \bm\theta) = \text{erf}(\theta^T \x)$ and choose the quadratic loss function $\ell(y,\, \hat{y})=\frac{1}{2}(y-\hat{y})^2$. This corresponds to a single hidden-layered neural network with non-linearity given by the error function $\text{erf}$. The choice of the error function allows a simple closed form expression for the functions $V(\bm\theta)$ and $U(\bm\theta,\,\bm\theta^\prime)$ introduced in section \ref{ssubsec:square_loss}, viz.,
\begin{equation}
\begin{aligned}
&\nabla V(\bm\theta) = -\frac{b(\bm\theta)^{-1/2}}{3\sqrt{\pi}} \sum_{i=1}^{2}y_i \sum_{j=1}^{3} a_{ij}(\bm\theta)\hat{\mathbf m}_i^j \\
\end{aligned}
\end{equation}
where $b(\bm\theta):=1+2s^2\|\bm\theta\|_2^2$,  and $y_1 = +1$, $y_2 = -1$ as class labels and 
\begin{equation}
\begin{aligned}
    &a_{ij}(\bm\theta) := e^{\frac{( \bm\theta^T \mathbf m_i^j)^2}{b(\bm\theta)} } \\
    &\hat{\mathbf m}_i^j := \mathbf m_i^j -\frac{2s^2 \bm\theta^T \mathbf m_i^j}{b(\bm\theta)}\bm\theta
\end{aligned}
\end{equation}
We also obtained a somewhat similar expression for $\nabla_1 U(\bm\theta, \bm\theta^\prime)$. 

\begin{figure*}[htbp]
\begin{center}
\resizebox{1\textwidth}{!}{%
\includegraphics[height=4cm]{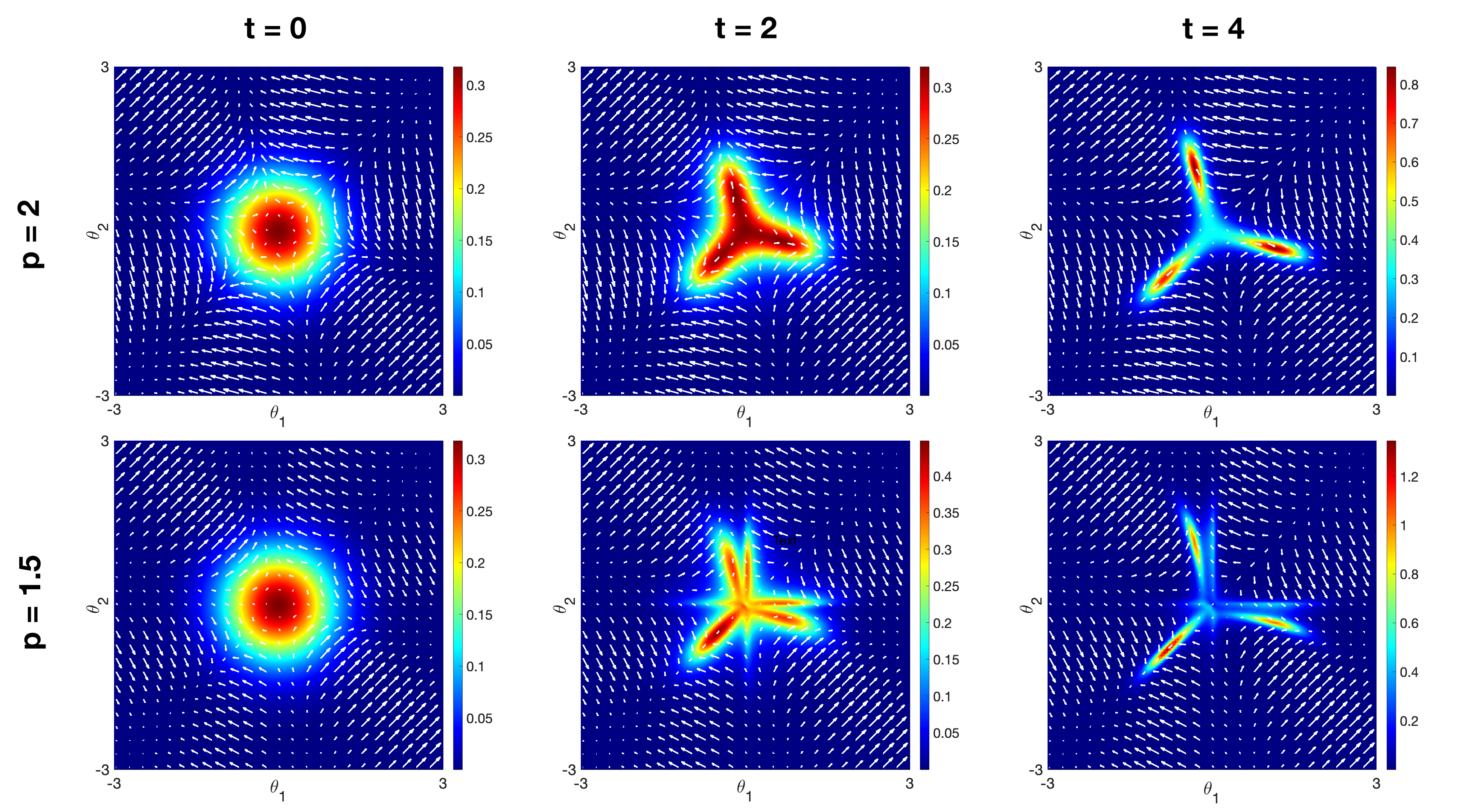}%
\quad
\includegraphics[height=4cm]{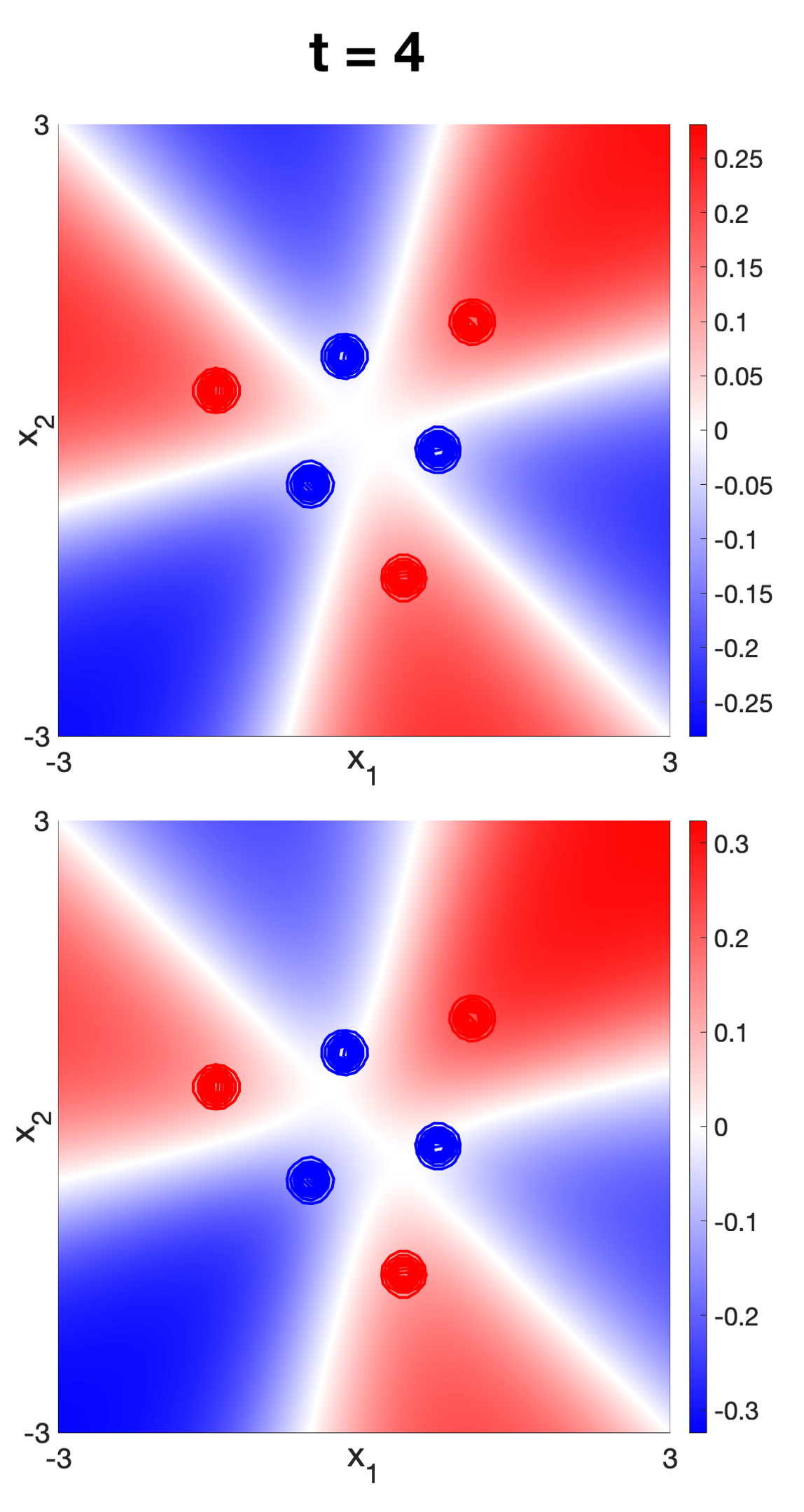}%
}
\end{center}
\caption{The results of solving the PDE (\ref{eq:SGD_PDE}). The top row corresponds to mirror potential $\|\cdot\|_2^2$, yielding SGD, and the bottom row corresponds to mirror potential $\|\cdot\|_{1.5}^2$. The first three columns correspond to the distribution of the two-dimensional parameter vector at times $t=0,2,4$, where $t=0$ represents the beginning of training and $t=4$ the end of training. The fourth column represents the output of the network for various inputs after training has ended ($t=4$). The expected risk at the completion of training is 0.4136 for the $\ell_2$ norm and 0.4118 for the $\ell_{1.5}$ norm. }
\label{fig:solvePDE}
\end{figure*}

Finally, we considered the SMD algorithm for this data distribution for two different mirror functions, $\|\cdot\|_p^2$ for $p=2$ and $p=1.5$. Note that $p=2$ gives us the standard SGD. We solved the PDE (\ref{eq:SGD_PDE}) for these two mirrors and show the results in Figure \ref{fig:solvePDE}. 

 The top row corresponds to mirror potential $\|\cdot\|_2^2$, and the bottom row corresponds to mirror potential $\|\cdot\|_{1.5}^2$. The first three columns correspond to the distribution of the two-dimensional parameter vector at times $t=0,2,4$, where $t=0$ represents the beginning of training and $t=4$ the end of training. For both SMDs the two-dimensional weight vector was initialized with a zero-mean Gaussian distribution. The evolution of the distribution of the weights can be seen at times $t=2$ and $t=4$. In the case of SGD, the weights appear to converge to three point masses that represent the three lines that separate the centers of the two data classes. In the case of $p=1.5$, the results are more interesting. There appears to be more than three types of nonlinear units; in fact, there is mass along the $\theta_1$ and $\theta_2$ axes, signifying that some of fraction of the weights are zero. This seems reasonable behavior since compared to $p=2$, the mirror potential with $p=1.5$ will encourage sparsity---since it has implicit bias towards minimizing the $\ell_{1.5}$ norm \cite{azizan_stochastic_2019,gunasekar_characterizing_2020}. 
 
 The fourth column represents the output of the network for various inputs after training has ended ($t=4$). For $p=2$,  we see that the two classes have been separated by three lines passing through the origin. For $p=1.5$, the structure of the output is more interesting. The classes are separated by more than three line segments, and not all go through the origin. Clearly, the two mirrors have solved the classification problem in two different ways.
 
 Despite this, the end performance of both trained models is nearly identical: the expected risk at the completion of training is 0.4136 for the $\ell_2$ norm and 0.4118 for the $\ell_{1.5}$ norm. Interestingly, $p=1.5$ performs slightly better.

\section{Outline of the Proof}\label{sec:proof}
Here we give a brief outline of the proof of Theorem~\ref{thm:main}. The techniques we use are similar to those of \cite{wang2017scaling} and \cite{sirignano_mean_2020}. \\
We start by introducing equivalent mirror domain counterparts of SMD iterates at \eqref{eq:SMD_updates_1} and proceed with analyzing the exchangeable Markov process in the mirror domain. For  $\bm\theta\in\R^d$, $\bm\omega := \nabla\psi(\bm\theta)$ denotes its mirror domain counterpart, and $\pi := \rho \circ \nabla\psi^{-1}$ denotes probability measure in the mirror domain\footnote{For a probability measure $\rho$ on a set $\S$ and a function $f:\S \rightarrow \S$, the push-forward of $\rho$ under $f$ is defined as $(\rho \circ f^{-1})(B) := \rho(f^{-1}(B))$ for all Borel $B\subset \S$.}. We rewrite SMD iterations with this new set of notations as follows
\begin{equation}\label{eq:mirror_domain_SMD}
    \begin{cases}
    \hat{\pi}_n^k = \frac{1}{n} \sum_{i=1}^{n} \delta_{\bm\omega_i^k}~,\\
    \bm\omega_i^{k+1} = \bm\omega_i^{k} + \frac{\tau}{n} \mathbf G(\bm\omega_i^k, \hat{\pi}_n^k, \mathbf z^{k+1})~,
    \end{cases}
\end{equation}
where $\mathbf G(\bm\omega, \pi, \mathbf z) := \mathbf F(\nabla \psi^{-1}(\bm\omega), \pi \circ \nabla\psi, \mathbf z)$. \\ 
We continue by characterizing the asymptotic limit of mirror domain empirical distributions $\{\hat{\pi}_n^k\}_{k=0}^{K}$, and consequently convert the results to the original parameter domain by a change of variables. To characterize the evolution of the empirical measure of parameters, we analyze the change in integral of a test function. Let $f \in \C_b^2(\R^d)$, and consider the the following consecutive difference,
\begin{equation}
\label{eq:difference}
    \begin{aligned}
    \langle &\hat{\pi}_n^{k+1}, f \rangle -  \langle \hat{\pi}_n^{k}, f \rangle =\frac{1}{n}\sum_{i=1}^{n} \left[f(\bm\omega_i^{k+1}) - f(\bm\omega_i^{k}) \right],\\
    &=\frac{1}{n}\sum_{i=1}^{n}\left[\frac{\tau}{n} \nabla  f(\bm\omega_i^{k})^T \mathbf G(\bm\omega_i^k, \hat{\bm\pi}_n^k, \mathbf z^{k+1}) +  r_i^k\right],
    \end{aligned}
\end{equation}
where the last equality is simply the first-order Taylor expansion and $r_i^k$ is the remainder term which is given as,
\begin{equation}
    r_i^k = \frac{1}{2}(\bm\omega_i^{k+1}-\bm\omega_i^k)^ T \nabla^2 f(\bar{\bm\omega}_i^k)(\bm\omega_i^{k+1}-\bm\omega_i^k) 
\end{equation}
for a point $\bar{\bm\omega}_i^k$ that lies on the line segment $[\bm\omega_i^k, \bm\omega_i^{k+1}]$. Proceeding on-wards,  we introduce the following three terms,
\begin{align}
    d^k &:= \frac{\tau}{n^2}\sum_{i=1}^{n} \nabla  f(\bm\omega_i^{k})^T \mathbb E_{\z \sim \mu(d\mathbf z)}\left[\mathbf G(\bm\omega_i^k, \hat{\pi}_n^k, \mathbf z)\right]  \\
    m^k &:= \frac{\tau}{n^2}\sum_{i=1}^{n} \nabla f(\bm\omega_i^{k})^T \mathbf G(\bm\omega_i^k, \hat{\pi}_n^k, \mathbf z^{k+1}) - d^k \\
    r^k &:= \frac{1}{n}\sum_{i=1}^{n} r_i^k
\end{align}
Rewriting~\eqref{eq:difference} in terms of these variables gives,
\begin{equation}
    \langle \hat{\pi}_n^{k+1}, f \rangle -  \langle \hat{\pi}_n^{k}, f \rangle = d^k + m^k + r^k.
    \label{eq:signle_difference}
\end{equation} 
Notice that using the definition of the drift function~\eqref{eq:velocity_definition}, $\mathbb E_{\z \sim \mu(d\mathbf z)}\left[\mathbf G(\bm\omega, \hat{\pi}, \mathbf z)\right]$ can be written as,
\begin{equation}
     \mathbb E_{\z \sim \mu(d\mathbf z)}\left[\mathbf G(\bm\omega, \hat{\pi}, \mathbf z)\right] = \v (\nabla\psi^{-1}(\bm\omega), \, \hat\pi \circ \nabla\psi ).
\end{equation}
We use the shorthand notation $\mathbf u (\bm\omega, \, \pi) :=  \v (\nabla\psi^{-1}(\bm\omega), \,\pi \circ \nabla\psi )$ to denote the mirror domain counterpart of $\v$. Consider a continuous-time process $\{\bar\pi_n(t)\, : \, t \in [0,T]\}$ similar to our earlier definition in~\eqref{eq:CT_embedding}. Taking the sum of~\eqref{eq:signle_difference} for $0\leq k < \lfloor{nt/\tau}\rfloor$,
\begin{equation}
\label{eq:sum_differences}
    \begin{aligned}
    \langle \bar{\pi}_{n}(t), f \rangle &- \langle \hat{\pi}^0, f \rangle = \sum_{k=0}^{\lfloor{nt/\tau}\rfloor -1} \left(\langle \hat{\pi}_n^{k+1}, f \rangle - \langle \hat{\pi}_n^{k}, f \rangle \right) \\
    &=\sum_{k=0}^{\lfloor{nt/\tau}\rfloor -1} d^k + \sum_{k=0}^{\lfloor{nt/\tau}\rfloor -1}m^k + \sum_{k=0}^{\lfloor{nt/\tau}\rfloor -1}r^k.
    \end{aligned}
\end{equation} 
The first sum on the RHS can be written as,
\begin{equation}
    \label{eq:first_sum}
    \begin{aligned}
    \sum_{k=0}^{\lfloor{nt/\tau}\rfloor -1} d^k 
    &= \frac{\tau}{n}\sum_{l=0}^{\lfloor{nt/\tau}\rfloor -1}  \langle \hat{\pi}_n^k, \nabla f(\cdot)^T \mathbf u(\cdot,\, \hat{\pi}_n^k) \rangle, \\
    &= \int_{0}^{[\,t\,]_n} \langle \bar{\pi}_{n}(s), \, \nabla f(\cdot)^T \mathbf u(\cdot,\bar{\pi}_{n}(s)) \rangle ds.
    \end{aligned} 
\end{equation}
where $[\,t\,]_n := \lfloor{\frac{nt}{\tau}}\rfloor \frac{\tau}{n}$, and the last equality derived from the definition of the process $\bar\pi_n(t)$. By replacing~\eqref{eq:first_sum} in~\eqref{eq:sum_differences} we have,
\begin{equation}
\begin{aligned}
    \langle \bar{\pi}_{n}(t),\, f \rangle = \langle\hat{\pi}^0,\,f \rangle &+ \int_{0}^{[\,t\,]_n} \langle\bar{\pi}_{n}(s),\,\nabla f(\cdot)^T \mathbf u(\cdot,\bar{\pi}_{n}(s)) \rangle ds \\
         &+\sum_{l=0}^{\lfloor{nt/\tau}\rfloor -1}m^l + \sum_{l=0}^{\lfloor{nt/\tau}\rfloor -1}r^l.
    \label{eq:sum_replaced}
\end{aligned}
\end{equation}
The proof concludes by showing the two sums on the RHS of~\eqref{eq:sum_replaced} converge to zero as $n\rightarrow \infty$ and $\frac{\tau}{n}=O(1/n)$. The convergence of the first sum comes from the fact that $\{m^l\}_{\ell>=0}$ can be viewed as deviations of a martingale form its mean, and therefore can be bounded by martingale concentration inequalities (e.g. see Lemma A.1 in~\cite{mei_mean_2018}). The last sum also approaches to zero due to our assumptions on the boundedness of the second derivatives of $\sigma$ and $\ell$. Thus, we have the following:  
\begin{equation}
    \langle\pi(t),\,f \rangle = \langle\pi^0,\,f \rangle + \int_{0}^{t} \langle\pi(s),\,\nabla f(\cdot)^T \mathbf u(\cdot\pi(s)) \rangle ds .
\end{equation}
The result obtained above can be rewritten in terms of primal domain variables, namely $\v$, and $\rho(t)$ by performing a change of variables as defined earlier in this section. Then, Hessian of the mirror function comes as the Jacobian of this change of variables process.

\section{Conclusion}\label{sec:conc}
In this paper we explored the performance of mirror descent iterates through a mean-field lens for ensemble average models. We characterized the time evolution of the distribution of the parameters through a nonlinear PDE and gave a Riemannian interpretation. It would be interesting to further study the effect of the mirror potential on the performance of SMD-trained models for different learning problems via analytic and numerical study of this PDE. It would also be interesting to generalize these results to other network models and learning algorithms.

\bibliography{references}

\end{document}